\documentclass[runningheads]{llncs}
\usepackage{graphicx}
\usepackage{amsmath,amssymb} 
\usepackage{color}
\usepackage[width=122mm,left=12mm,paperwidth=146mm,height=193mm,top=12mm,paperheight=217mm]{geometry}
\usepackage{cite}
\usepackage{color, colortbl}
\definecolor{Gray}{gray}{0.9}

\usepackage{times}
\usepackage{booktabs}
\usepackage{multirow}
\usepackage{subfigure}

\usepackage[font={small}]{caption}
\begin{document}

\pagestyle{headings}
\mainmatter





\title{Dendritic Spine Shape Analysis: A Clustering Perspective}

\titlerunning{Dendritic Spine Shape Analysis: A Clustering Perspective}  
 \toctitle{Dendritic Spine Shape Analysis: A Clustering Perspective}
\author{Muhammad Usman Ghani \inst{1} \and Ertun\c c Erdil \inst{1} \and S\"{u}meyra Demir Kan{\i}k \inst{1} \and Ali \"Ozg\"ur Argun\c sah \inst{2} \and Anna Felicity Hobbiss \inst{2} \and Inbal Israely \inst{2} \and Devrim \"{U}nay \inst{3} \and Tolga Ta\c sdizen \inst{4} \and M\"{u}jdat \c Cetin \inst{1}}
\authorrunning{Ghani et al.} 
%
\tocauthor{Muhammad Usman Ghani, Ertun\c c Erdil, S\"{u}meyra Demir Kan{\i}k, Ali \"Ozg\"ur Argun\c sah, Anna Felicity Hobbiss, Inbal Israely, Devrim \"{U}nay, Tolga Ta\c sdizen, M\"{u}jdat \c Cetin}
\institute{Faculty of Engineering and Natural Sciences, Sabanci University, Istanbul, Turkey.\\
\email{\{ghani,sumeyrakanik,mcetin\}@sabanciuniv.edu}
\and
Champalimaud Neuroscience Programme, Champalimaud Centre for the Unknown, Lisbon, Portugal.\\
\email{\{ali.argunsah,anna.hobbiss,inbal.israely\}@neuro.fchampalimaud.org}
\and
Faculty of Engineering and Computer Sciences, Izmir University of Economics, Izmir, Turkey.\\
\email{devrim.unay@ieu.edu.tr}
\and
Electrical and Computer Engineering Department, University of Utah, USA.\\
\email{tolga@sci.utah.edu}
}

\maketitle

\begin{abstract}
Functional properties of neurons are strongly coupled with their morphology. Changes in neuronal activity alter morphological characteristics of dendritic spines.
First step towards 
understanding the structure-function relationship is to group spines into main spine classes reported in the literature. Shape analysis of dendritic spines can help neuroscientists understand the underlying relationships. Due to unavailability of reliable automated tools, this analysis is currently performed manually which is a time-intensive and subjective task. Several studies on spine shape classification have been reported in the literature, however, there is an on-going debate on whether distinct spine shape classes exist or whether spines should be modeled through a continuum of shape variations. Another challenge is the subjectivity and bias that is introduced due to the supervised nature of classification approaches. In this paper, we aim to address these issues by presenting a clustering perspective. In this context, clustering may serve both confirmation of known patterns and discovery of new ones. We perform cluster analysis on two-photon microscopic images of spines using morphological, shape, and appearance based features and gain insights into the spine shape analysis problem. We use histogram of oriented gradients (HOG), disjunctive normal shape models (DNSM), morphological features, and intensity profile based features for cluster analysis. We use x-means to perform cluster analysis that selects the number of clusters automatically using the Bayesian information criterion (BIC). For all features, this analysis produces $4$ clusters and we observe the formation of at least one cluster consisting of spines which are difficult to be assigned to a known class. This observation supports the argument of intermediate shape types.
\end{abstract}

\section{Introduction}
Dendritic spines, small protrusions of the dendritic shaft, are one of the  most important structures of neurons. Ram\'{o}n y Cajal first identified spines in the $19$th century and suggested that neuronal activity variations change the spine morphology \cite{paper4,yuste2010dendritic}. This claim has been supported by several studies reporting changes in the morphology and density with changes in neuronal activity \cite{paper3, matsuzaki2004structural, harvey2007locally, govindarajan2011dendritic}. Spines are the post-synaptic partners of a synapse \cite{paper16} and are main receivers for synaptic input \cite{yuste2010dendritic}. Dendritic spines in the hippocampal neurons are related with learning and short-term memory \cite{paper13, xu2006optical}. Studies also reported that spine density is decreased due to some neuro-degenerative diseases such as Alzheimer's \cite{xu2006optical}. \par

\begin{figure}[tb]
	\centering
		\includegraphics[width=0.49\textwidth]{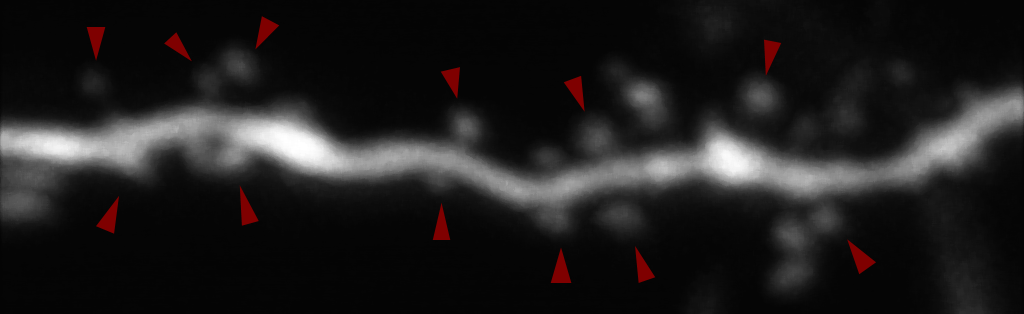}
	\caption{A dendritic branch with several spines imaged using a two-photon laser scanning microscope (2PLSM).}
	\label{fig:DendriticSpines}
\end{figure}

\begin{figure}[tb]
    \centering
    \subfigure[Intensity images collected using 2PLSM]{\includegraphics[width=0.3\textwidth]{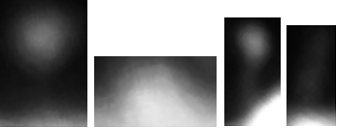}
}
\hspace{10pt} 
    \subfigure[Manual annotations]{\includegraphics[width=0.3\textwidth]{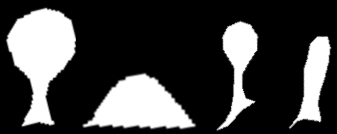}
        }
    \caption{Spine Classes: Mushroom, Stubby, Thin, Filopodia (Left to Right). Intensity and corresponding manually annotated images are shown for each shape class.}
    	\label{fig:SpineClasses}
\end{figure}

A dendritic branch with several spines is shown in Fig. \ref{fig:DendriticSpines}. Each spine has two segments, head and neck. Spine head is connected to the parent dendrite through the neck \cite{paper14}. 
Dendritic spines exhibit extraordinary diversity \cite{paper10}; they have different sizes and densities across different cell types, brain areas, and animal species \cite{yuste2010dendritic}. A great variety in spine head and neck dimensions is usually demonstrated even within the same cell \cite{yuste2010dendritic}. These facts emphasize the challenging nature of the spine analysis task.
Dendritic spines have different shape types; researchers suggest that different morphological variations could possibly be related to various developmental stages or functional roles \cite{parnass2000analysis}. In the literature, dendritic spines have mostly been grouped into four shape classes: filopodia, mushroom, thin, and stubby \cite{yuste2010dendritic, paper13, paper12, paper5, paper8}. Filopodia spines have long necks and no heads, mushroom spines have long necks and large bulbous heads, thin spines have long necks and small heads, and stubby spines are known to have either no necks or short necks \cite{yuste2010dendritic}. An example of each of these classes is given in Fig. \ref{fig:SpineClasses}. Distribution of different shape types varies in different areas of the brain; it is also dependent upon the age of the animal being imaged, for instance stubby spines are dominant during early postnatal development \cite{yuste2010dendritic}.\par

This classification of spine shapes has been widely used in the literature, however, there is an open research question concerning whether distinct classes of spines shapes exist or whether spines should be modeled through a continuum of shape variations. Parnass et al. \cite{parnass2000analysis} proposed that morphological groups of spine shapes do not represent inherent shape types, instead they depict shape variations a spine can go through during its life time. 
Bourne and Harris \cite{bourne2007thin} noticed spine enlargement as a result of synaptic enhancement, causing transition of thin spines to mushroom type. 
Peters and Kaiserman-Abramof \cite{paper8} reported the existence of spines with intermediate shape types and they found it difficult to assign them to one of the standard shape types. Basu et al. \cite{Basu01042016} reported a human expert being unsure while assigning labels to some of the spines. Arellano et al. \cite{paper7} who used morphological features for spine analysis, also found several spines with intermediate morphological characteristics in their dataset. Spacek and Hartman \cite{paper9} could not classify some spines into standard shape types and introduced a new class between mushroom and stubby, and thin and mushroom spines. Ruszczycki et al. \cite{paper10} hinted towards a different classification standard: classifying spines into large and small, they reported better sensitivity with this classification approach. Wallace and Bear \cite{WallaceBear04} used spine length and head diameter to perform spine analysis and found a continuous distribution. Mancuso et al. \cite{mancuso2013methods} suggested using morphological features to perform clustering and count spines in different clusters. 
In summary, different groups work with single or multiple neuroscience experts and each group uses their defined rules for classification, which results in subjectivity.\par

Quantitative analysis of dendritic spines is important for neurobiological research as it can help neuroscientists understand the underlying structure-function relationship. Currently this analysis is performed manually due to unavailability of reliable automated spine shape analysis tools. Manual analysis is a laborious, time-intensive, and most importantly subjective task. Rodriguez et al. \cite{paper12} reported inter-operator and intra-operator variations in the spine type labeling task. Availability of reliable automated analysis tools can expedite research in this domain and assist neuroscientists decode the underlying relationship between neuron function and structure.\par



One might question why perform clustering rather than treating this as a classification problem. First of all, classification methods use manually provided labels as ground truth and extracting those labels is a time-intensive task. It also introduces subjectivity, which could be reduced by employing several experts and using a majority vote approach but this would make the labeling effort even more time-intensive. Inter-operator and intra-operator variability reported by Rodriguez et al. \cite{paper12} emphasizes that subjectivity is a major issue in performing classification. Another issue with supervised classification is that it inherently starts from a pre-defined set of classes and does not allow exploration of potential intermediate shapes or possible continuous variation of shapes. Although clustering does not explicitly enable the latter either, it can be viewed as a step in that direction. Furthermore, some existing techniques require manual annotation of spines either to directly use them for feature extraction or for training segmentation algorithms. 
The objective of clustering in this context is two-fold: confirm the hypothesis of some distinct shape classes and discover new natural groups. We discover natural groups in the data using different features and analyze whether they support the existing hypotheses or add new information to our understanding of spine shapes. \par

As suggested by Mancuso et al. \cite{mancuso2013methods}, we present a clustering-based approach for spine shape analysis. We perform cluster analysis using several feature representations and gain insights by performing analysis of discovered natural groups. We use Disjunctive Normal Shape Models (DNSM) \cite{Messadi_MICCAI}, Histogram of Oriented Gradients (HOG) \cite{dalal2005histograms}, intensity profiles \cite{erdil2015joint}, and morphological features \cite{Ghani_SIU15}. We use an extension of k-means, x-means \cite{pelleg2000x}, to perform cluster analysis that uses the Bayesian Information Criterion (BIC) to select the number of clusters automatically. This study is based on two-photon laser scanning microscopy ($2$PLSM) images. Analyzing $2$PLSM images is more challenging in comparison to confocal laser scanning microscopy (CLSM) images due to low signal to noise characteristics. Additionally, following the Abbe's law \cite{lipson2010optical}, resolution of $2$PLSM images is half of the CLSM images. The reason behind using $2$PLSM is that it allows imaging of living cells, which would capture shape transitions during synaptic process \cite{paper14, so2000two}.


The major contribution of this paper is application of HOG-based features for spine analysis and cluster analysis of dendritic spines with different representations
. To the best of the authors' knowledge, this is the first paper that performs such an analysis of dendritic spine shapes with a wide range of feature sets.
The rest of this paper is structured as follows. A brief summary of some of the related work is presented in Section \ref{relWork}. Section \ref{meth} discusses the methodology of our approach in detail. Experimental analysis and results are presented and discussed in Section \ref{exp}. Section \ref{conc} summarizes the findings and conclusions of this paper.\par

\section{Related Work} \label{relWork}
There exist several studies on supervised spine classification but none of these studies have reported performing unsupervised cluster analysis of dendritic spine shapes. Rodriguez et al. \cite{paper12} performed spine classification on $3$D images using morphological features. They developed a decision tree based classifier and evaluated its performance using labels provided by human experts. Son et al. \cite{paper13} also developed a classification approach using morphological features and evaluated their approach with labels assigned by a human expert. Shi et al. \cite{paper16} developed a semi-supervised learning approach for spine classification based on morphological features, and used human experts for validation of their results. A recent study on spine analysis applied ISOMAP \cite{tenenbaum_global_2000} to study the importance of different morphological parameters and found neck length and head diameter to be the most prominent features for mushroom and stubby spines \cite{ghani2016manifold}. Ghani et al. \cite{ghani2016dnsm} exploited the parametric nature of the DNSM approach and used its parameters for spine classification; they also used labels assigned by a human expert for performance evaluation. Erdil et al. \cite{erdil2015joint} developed a joint classification and segmentation approach, within which they used intensity profiles for classification of spines. Labels assigned by a human expert were used to evaluate the performance of their algorithm.\par

As it can be noticed from a small subset of studies on classification summarized here, most of the groups use one or more human experts to assign class labels which are later used to evaluate the performance of their supervised classification approaches. Even though using the manually extracted labels as ground truth is a viable approach for this problem, it introduces subjectivity. We attempt to address this issue by presenting a clustering approach aiming to discover natural groups of spine shapes in an unsupervised fashion using various feature representations.\par  

\section{Methodology} \label{meth}
We provide the details of our methodology in this section. Post natal $7$ to $10$ days old mice are imaged using $2$PLSM.\footnote{All animal experiments are carried out in accordance with European Union regulations on animal care and use, and with the approval of the Portuguese Veterinary Authority (DGV).} We have acquired $15$ stacks of $3$D images using $2$PLSM. After applying median filtering, we project $3$D images to $2$D using maximum intensity projection (MIP) \cite{wallis1989three}. We used $2$D projections for this analysis, because resolution along the $z$-axis in our data is $0.3 \mu m$ which is much worse than lateral resolution, which is $0.019 \mu m$ or $0.024 \mu m$ for different stacks. The slices along $z$-axis provide limited information \cite{zhang2007dendritic}. The spines consist of a small head ($\sim 1 \mu m$ diameter) and a thin neck ($\sim 0.2 \mu m$ diameter), and are generally $0.5 \mu m$ to several $\mu m$ long \cite{yuste2010dendritic}. Due to low $2$PLSM resolution, complete spine covers only a few slices along the $z$-axis. The low axial resolution makes the spine analysis in $3$D very challenging even for human experts \cite{bai2007automatic}. While there are other projection methods available, MIP is a standard projection procedure used in most of the neuroscience studies \cite{paper14,bai2007automatic, xu2006shape, zhang2007dendritic,Basu01042016}. In total, $242$ dendritic spines have been selected from $15$ dendritic branches for this study.\par

\subsection{HOG features}
\begin{figure}[tb]
	\centering
		\includegraphics[width=0.45\textwidth]{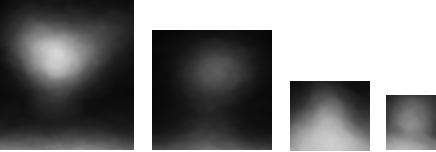}
	\caption{Sample images from the dataset prepared for HOG.}
	\label{fig:hog_dataset}
\end{figure}

HOG \cite{dalal2005histograms} computes histogram of gradient orientations and applies contrast normalization to improve performance.
It is observed that spine heads have uniform intensities whereas intensity in the neck region is not uniform. A decreasing intensity pattern can be noticed in the neck part of the spines. Using this appearance information would help us discover clusters with different appearance patterns. In order to compute HOG features, we select a region of interest (ROI) in intensity images such that the spine is completely inside the ROI. This does not require the ROI for all spines to have the same dimensions. Further, we rotate the ROI such that spine necks are vertically aligned. Examples resulting from this process are shown in Fig. \ref{fig:hog_dataset}.\par

In order to capture fair amount of small-scale details, we selected the cell size as a function of width and height: $CellSize=\big[height/5,width/5\big]$; cells are small spatial
regions. A large block size value allows to suppress local intensity changes; blocks are relatively large spatial regions. To keep moderate level of information about local illumination variations, we selected a block size value equal to twice the $CellSize$. Contrast normalization is controlled through block overlap, and we selected a block overlap of $1$ cell. We used $9$ signed histogram orientation bins, because using signed orientation allows to track light to dark and dark to light intensity changes. We computed $576$-dimensional HOG feature vectors with these settings and later used these features for cluster analysis.\par 

\subsection{DNSM features}

\begin{figure}[tb]
	\centering
		\includegraphics[width=0.4\textwidth]{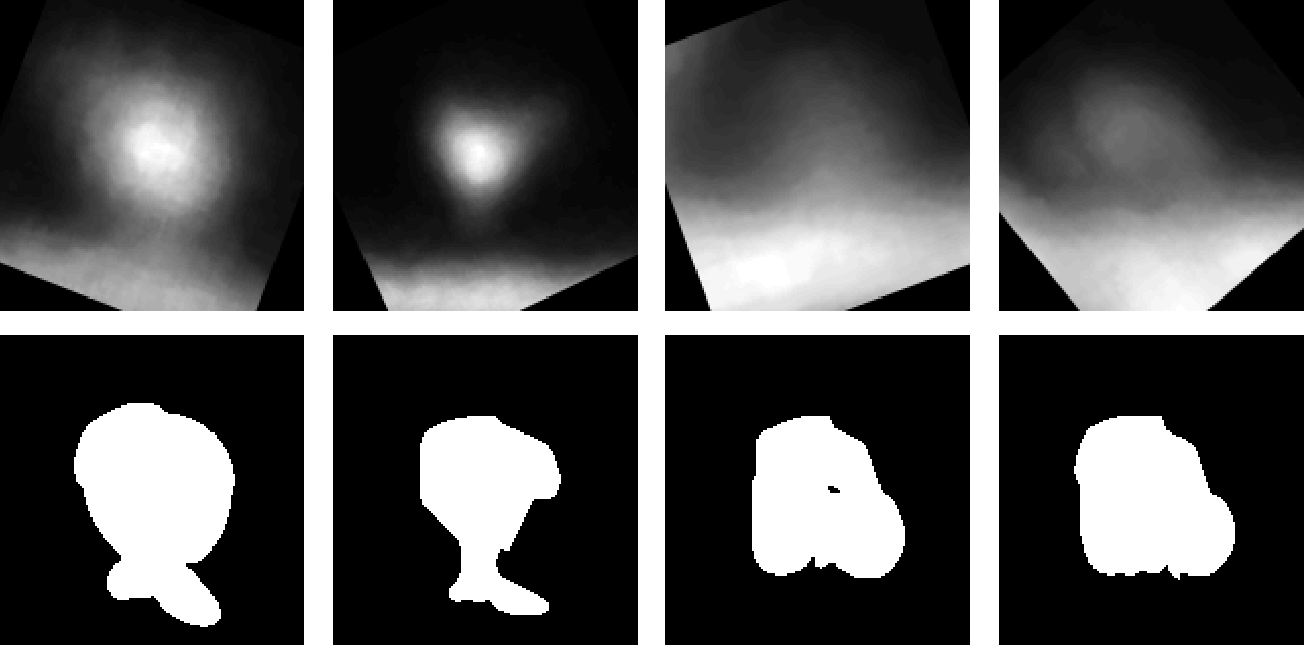}
	\caption{A few images from  dataset prepared for DNSM features: before segmentation (above) and segmented images (below).}
	\label{In_Seg}
\end{figure}

DNSM is a parametric shape model, proposed recently by Ramesh et al. \cite{Ramesh_ISBI}. DNSM represents a shape as a union of 
convex polytopes, which are constructed by intersections of half spaces. DNSM attempts to approximate the characteristic function of a shape
. For further details of the DNSM, readers are referred to \cite{Ramesh_ISBI,Messadi_MICCAI}.
DNSM-based features provide a shape representation; it would be an interesting experiment to perform clustering using DNSM-based shape features. We apply DNSM to segment dendritic spine images following the approach in \cite{ghani2016dnsm} and use $384$-dimensional DNSM parameters as  feature vectors to perform cluster analysis. A few images from the dataset used for DNSM features are presented in Fig. \ref{In_Seg}.\par


\subsection{Morphological features}
Morphology of dendritic spines has been extensively studied in the literature. Most of the studies on spine analysis compute morphological parameters to perform classification of spines. In this paper, we use $12$ morphological features suggested in a recent study on spine classification \cite{Ghani_SIU15}. The morphological features we use are listed in Table \ref{tab:morph_feat}. In order to compute these morphological features, we perform segmentation using DNSM and apply methods suggested in \cite{Ghani_SIU15}. 

\begin{table}[tb]
  \centering
  \caption{List of Morphological features used.}
    \begin{tabular}{|l|}
    \hline
    \rowcolor{Gray}
    Neck Length, Head Diameter, Circularity, Shape Factor \\ 
    Width and Height of bounding box, Perimeter, Area \\ 
    \rowcolor{Gray}
    Neck Length to Head Diameter Ratio (NHR) \\
    Foreground to background pixels ratio in bounding box \\
    \hline
    \end{tabular}%
  \label{tab:morph_feat}%
\end{table}%

\subsection{Intensity profiles based features}
\begin{figure}[tb]
\centering
\subfigure[First region]{
   \includegraphics[scale=0.4] {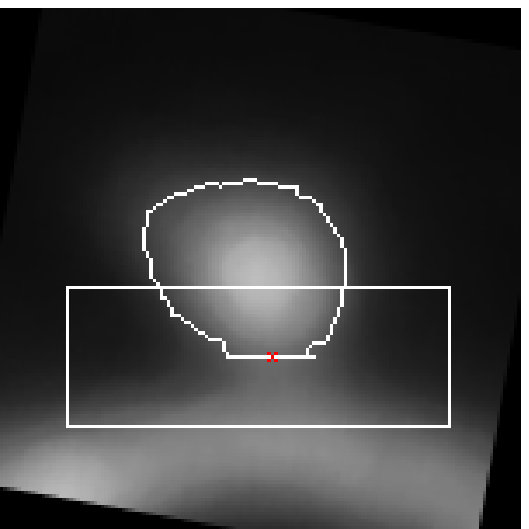}
   \label{fig:isbi15regions_1}
 }
 \hspace{50pt}
\subfigure[Second region]{
   \includegraphics[scale=0.4] {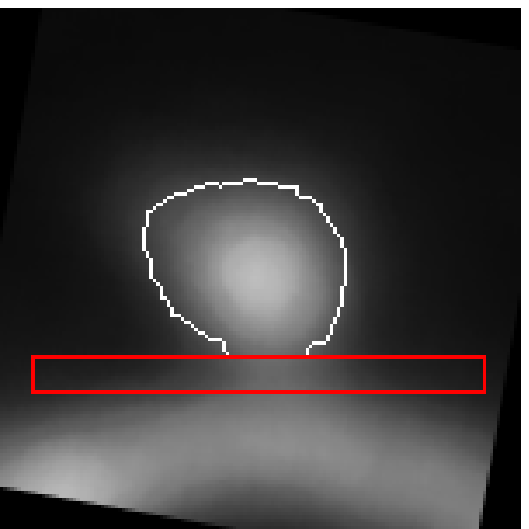}
   \label{fig:isbi15regions_2}
 }
\caption{Regions in which a potential neck is likely to be contained.}
\label{fig:isbi15regions}
\end{figure}

Erdil et al. \cite{erdil2015joint} suggests that intensity information in the regions in which a potential neck is likely to be contained can be used to differentiate spine classes. Regions where the neck might appear is found using the assumption that the spine neck lies below the spine head. Once the spine head is found by minimizing an intensity-based energy function using active contours \cite{chan2001active}, the approach in \cite{erdil2015joint} creates two rectangular regions below the spine head as shown in Fig. \ref{fig:isbi15regions}. The first region shown in Fig. \ref{fig:isbi15regions_1} is constructed such that the bottom point of the spine head (shown by a red cross) lies at the center of the rectangle. The second rectangular region shown in Fig. \ref{fig:isbi15regions_2} is a narrower one and is drawn such that it is located just below the spine head. Erdil et al. \cite{erdil2015joint} extract three sets of feature vectors by exploiting intensities in these rectangular regions which are combined to form $378$-dimensional feature vectors. The first set of feature vectors is obtained by summing up the intensities in the first rectangle horizontally. Similarly, the second set of feature vectors are obtained by vertical summation of the intensities in the corresponding rectangle. The final set of feature vectors are the histograms of intensities in the second rectangular region.

\subsection{Feature Selection}
Considering the high-dimensionality of feature representations being used (except morphological features), we apply a feature similarity based unsupervised feature selection algorithm \cite{mitra2002unsupervised}. Mitra et al. \cite{mitra2002unsupervised} introduced the maximum information compression index, which attempts to minimize the information loss while selecting a certain number of features. Here, the aim of feature selection is to aid the clustering algorithm, we select $100$ features for each feature representation (except morphological features) and use these selected features to perform clustering.\par

\subsection{Clustering}
Jain \cite{jain2010data} suggests there are two objectives for clustering: (i) exploratory: when there is no existing hypothesis or model, the aim is to discover patterns, and (ii) confirmatory: when a pre-specified model or hypothesis exists, the objective of cluster analysis is to confirm the model on the dataset being used. For dendritic spine analysis, the literature provides a pre-specified model as described in the introduction section. The nature of our analysis is: (i) an attempt to analyze how well a pre-specified model fits our data, (ii) if such a model does not fit our data, discover and explore natural groups within the data. \par

Jain \cite{jain2010data} argues that there is no best clustering algorithm, because every clustering technique implicitly or explicitly imposes a structure on the data, and it gives good results if there is a good match. Jain further emphasizes that it is rather crucial to select the appropriate representation that implicitly or explicitly makes the pattern discovery an easy process. Considering the clustering analysis problem as a selection of appropriate representation rather than  selection of a clustering method, we have compared different feature representations in terms of clustering results. We applied x-means \cite{pelleg2000x}, an extended version of k-means, which does not require the number of clusters to be provided. It uses BIC to automatically select the number of clusters in the available data from a given range of number of clusters, which we set as $2$ to $10$. It begins with lower bound of given range for number of clusters and continues computing clusters until upper bound for number of clusters have been reached; during this process it also computes BIC score for each cluster assignment. Finally, it selects the number of clusters based on best BIC score.\par
 
\section{Results and Discussion} \label{exp}
Our dataset consists of $242$ dendritic spines selected from $15$ dendritic branches for this analysis. These are spines that have been labeled as mushroom or stubby by a human expert. 
Analysis of clusters formed using different feature representations is presented in this section. \par

\subsection{HOG features based analysis}
\begin{figure}[tb]
	\centering
		\includegraphics[width=0.35\textwidth]{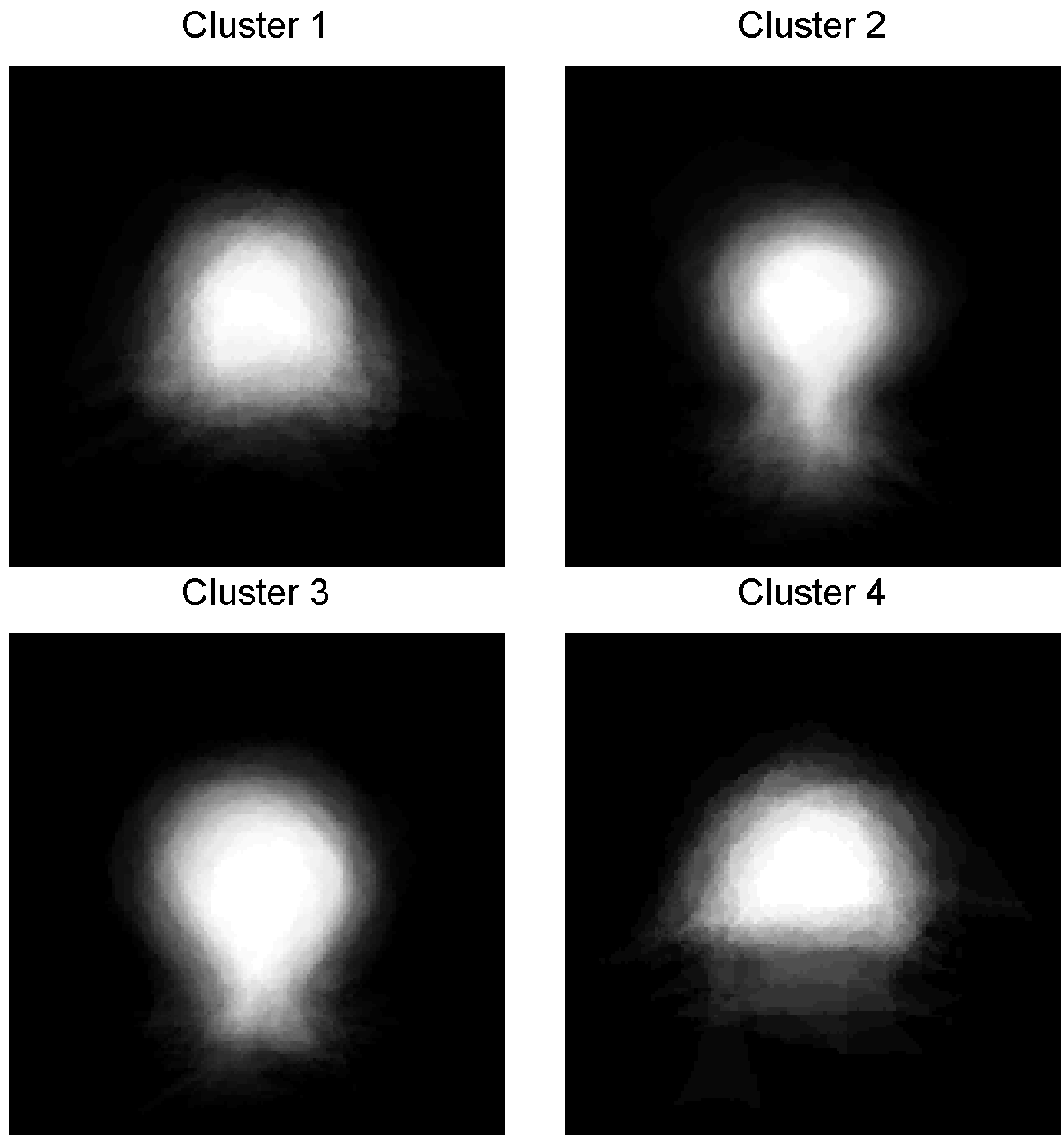}
	\caption{Average image for each cluster generated using the HOG features.}
	\label{fig:cl_hog}
\end{figure}

\begin{figure}[tb]
	\centering
\subfigure[Cluster 1]{
   \includegraphics[width=0.18\textwidth] {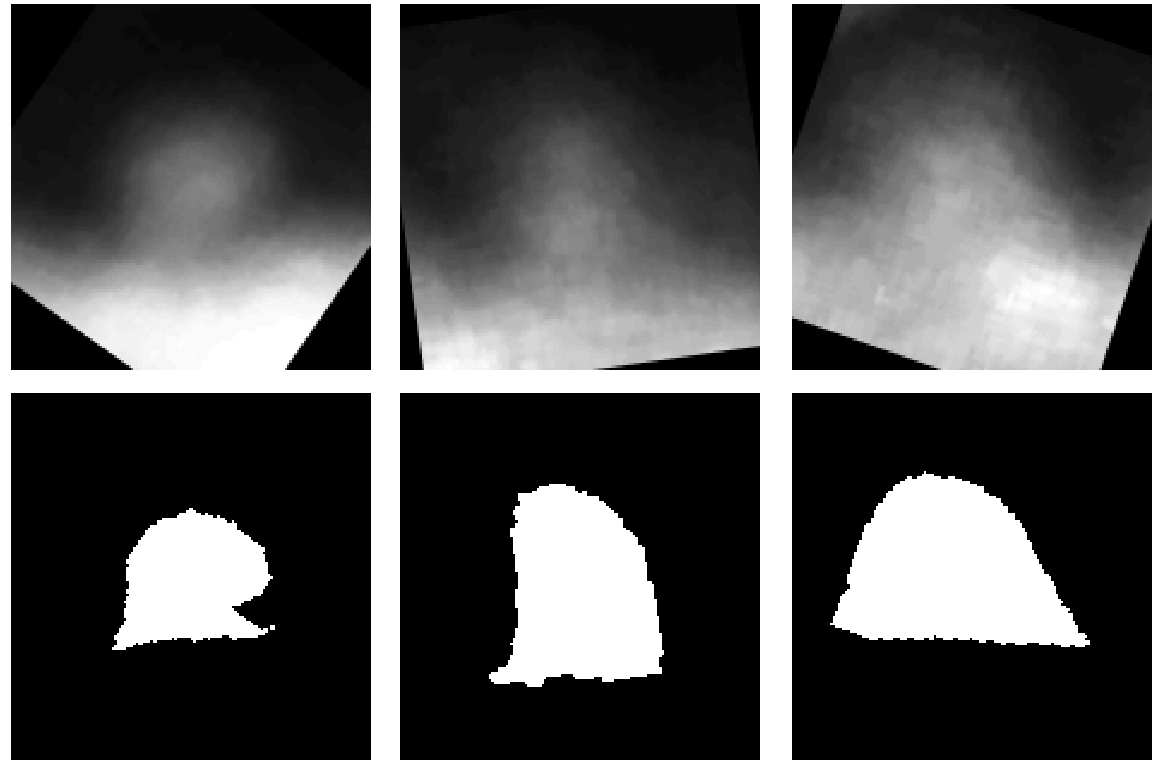}
   \label{fig:cl1_hog}
 }
 \hspace{30pt}
\subfigure[Cluster 4]{
   \includegraphics[width=0.18\textwidth] {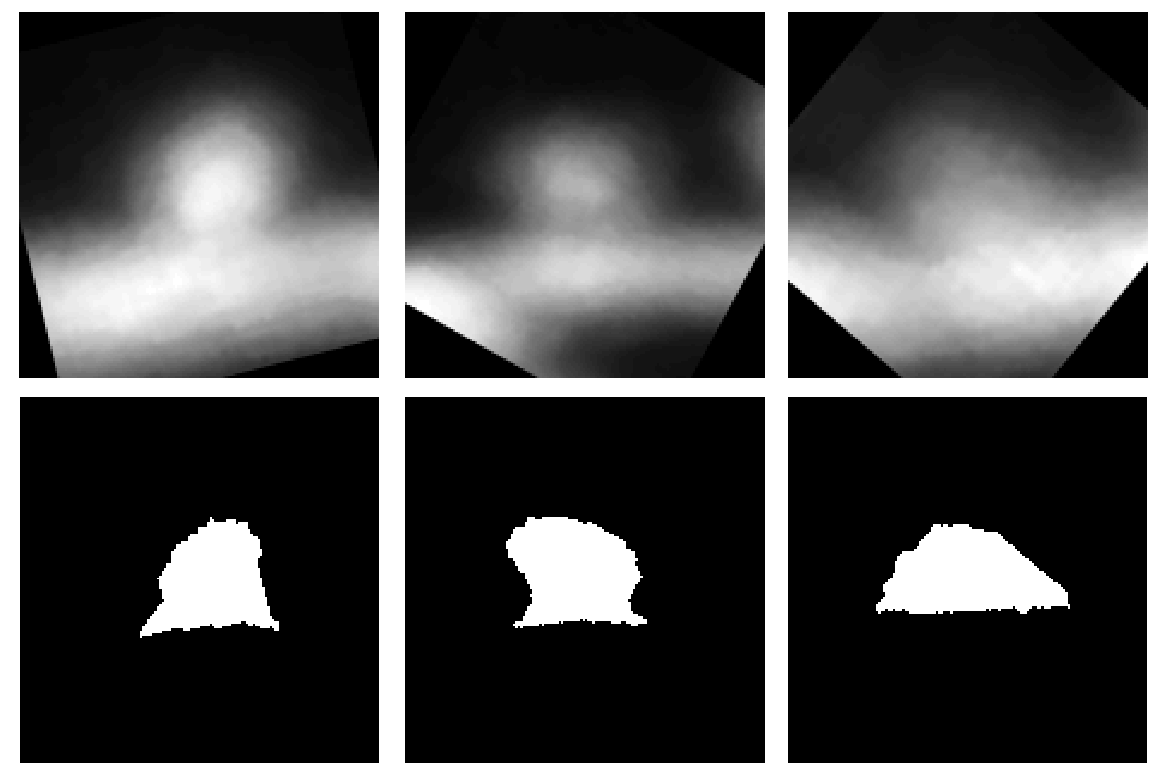}
   \label{fig:cl4_hog}
 }
	\caption{Intensity (top) and corresponding manually annotated images (bottom) for some of the spines grouped in cluster $1$ and cluster $4$ using the HOG features.}
	\label{fig:cl1_4_hog}
\end{figure}

Using HOG based appearance feature representation for x-means clustering resulted in $4$ clusters. The average image for each cluster is computed by averaging manually segmented binary images in that cluster. The resulting images are shown in Fig. \ref{fig:cl_hog}. There are $49$ spines in cluster $1$, $93$ spines in cluster $2$, $72$ spines in cluster $3$, and $28$ spines in cluster $4$. As it is evident from the average images, cluster $2$ and cluster $3$ represent mushroom spines (long neck and big head). However, clusters $1$ and $4$ appear to consist either of spines from both classes or of spines that may possibly lie in between these two classes in the shape space. When we examine individual samples from these clusters, illustrated in Fig. \ref{fig:cl1_4_hog}, we observe that they exhibit similar characteristics, i.e., have small heads and no necks. However, closer analysis of intensity images shows existence of short necks, i.e., low intensity regions just below the head part. These observations support the produced clusters in the sense that although there are some spines which are easy to be classified (grouped in clusters $2$ and $3$), even a human expert would have difficult time providing labels for most of the spines in cluster $1$ and cluster $4$. This analysis also points to what one might call two subclasses (cluster $2$ and cluster $3$) within the mushroom class.\par

\subsection{DNSM features based analysis}
\begin{figure}[tb]
	\centering
		\includegraphics[width=0.35\textwidth]{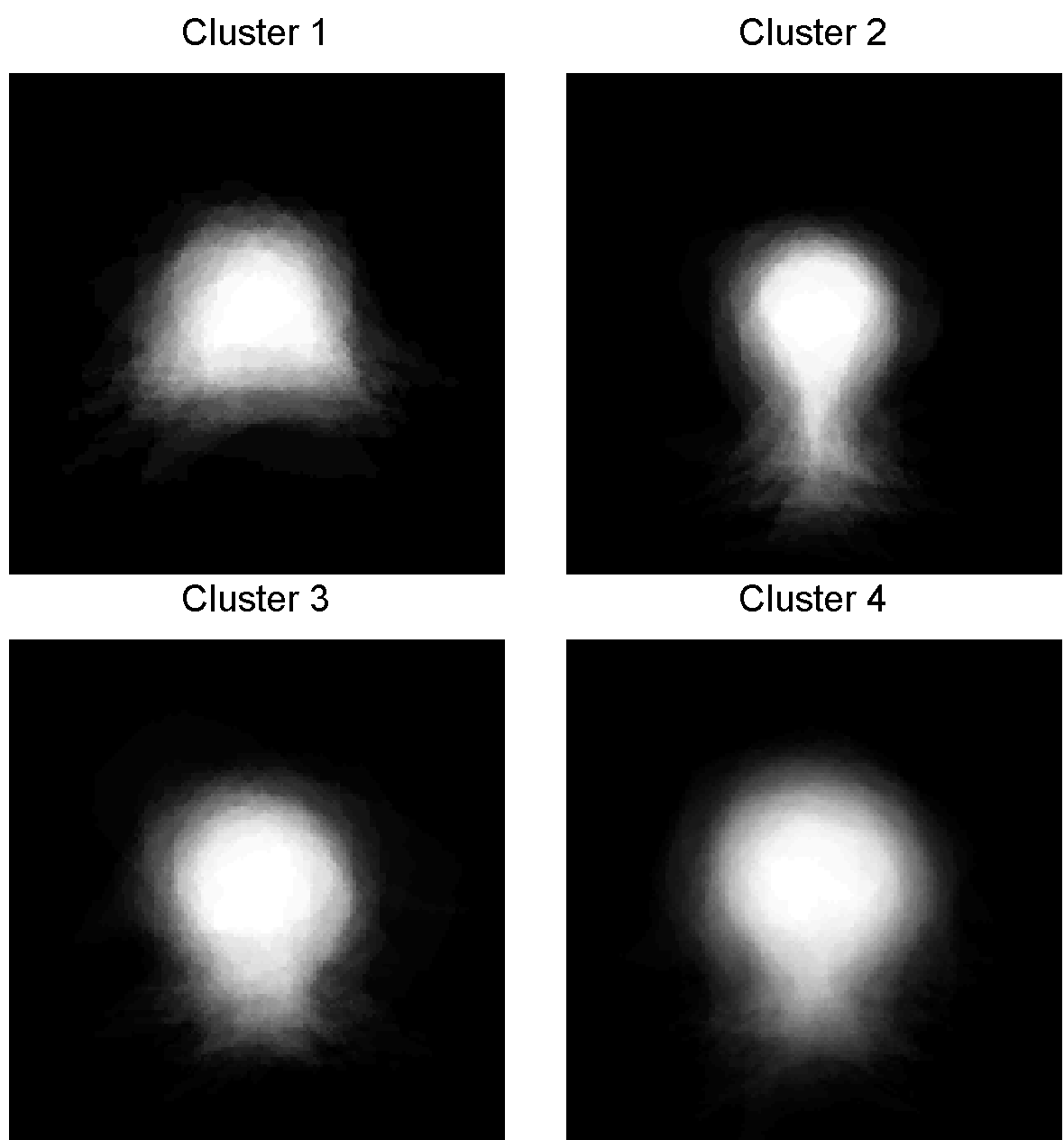}
	\caption{Average image for each cluster generated using the DNSM features.}
	\label{fig:cl_dnsm}
\end{figure}

\begin{figure}[tb]
	\centering
		\includegraphics[width=0.5\textwidth]{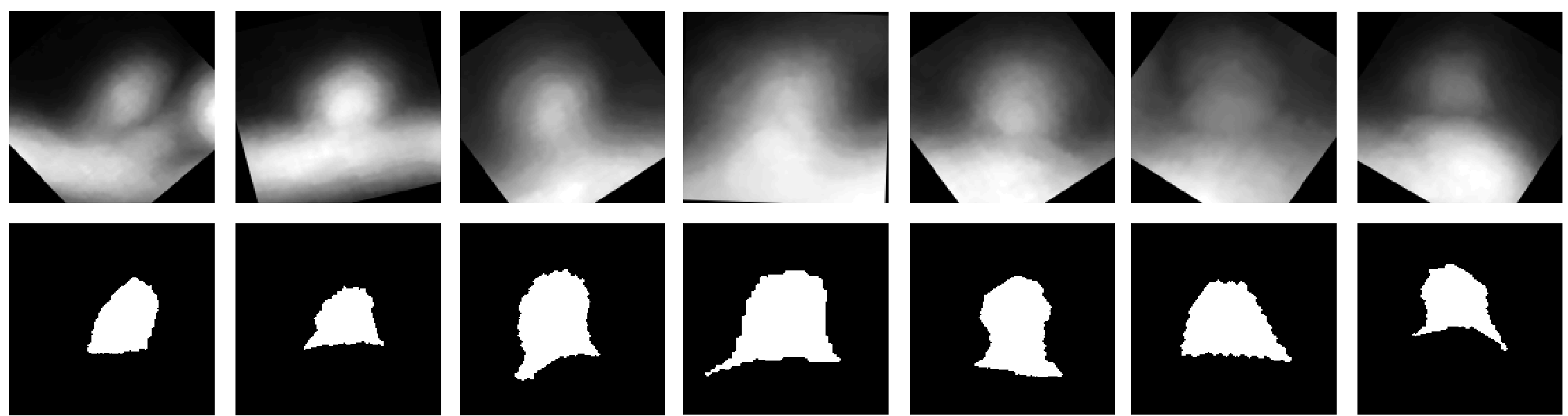}
	\caption{Intensity (top) and corresponding manually annotated images (bottom) for some of the spines grouped in cluster $1$  using the DNSM representation.}
	\label{fig:cl1_dnsm}
\end{figure}

We computed shape features using DNSM and performed clustering on this representation. The algorithm produced $4$ clusters consisting of $32$, $48$, $50$, and $112$ spines. Average images of these clusters are given in Fig. \ref{fig:cl_dnsm}. 
Most of the spines in cluster $1$ have short or no necks; their head diameter to neck diameter ratio is approximately $1$. A few spines from cluster $1$ are presented in Fig. \ref{fig:cl1_dnsm}. This cluster appears to contain spines that clearly exhibit the characteristics of stubby spines as well as spines with distinct heads and thick necks. Cluster $2$, cluster $3$, and cluster $4$ are mostly mushroom clusters. 

\subsection{Morphological features based analysis}
\begin{figure}[tb]
	\centering
		\includegraphics[width=0.35\textwidth]{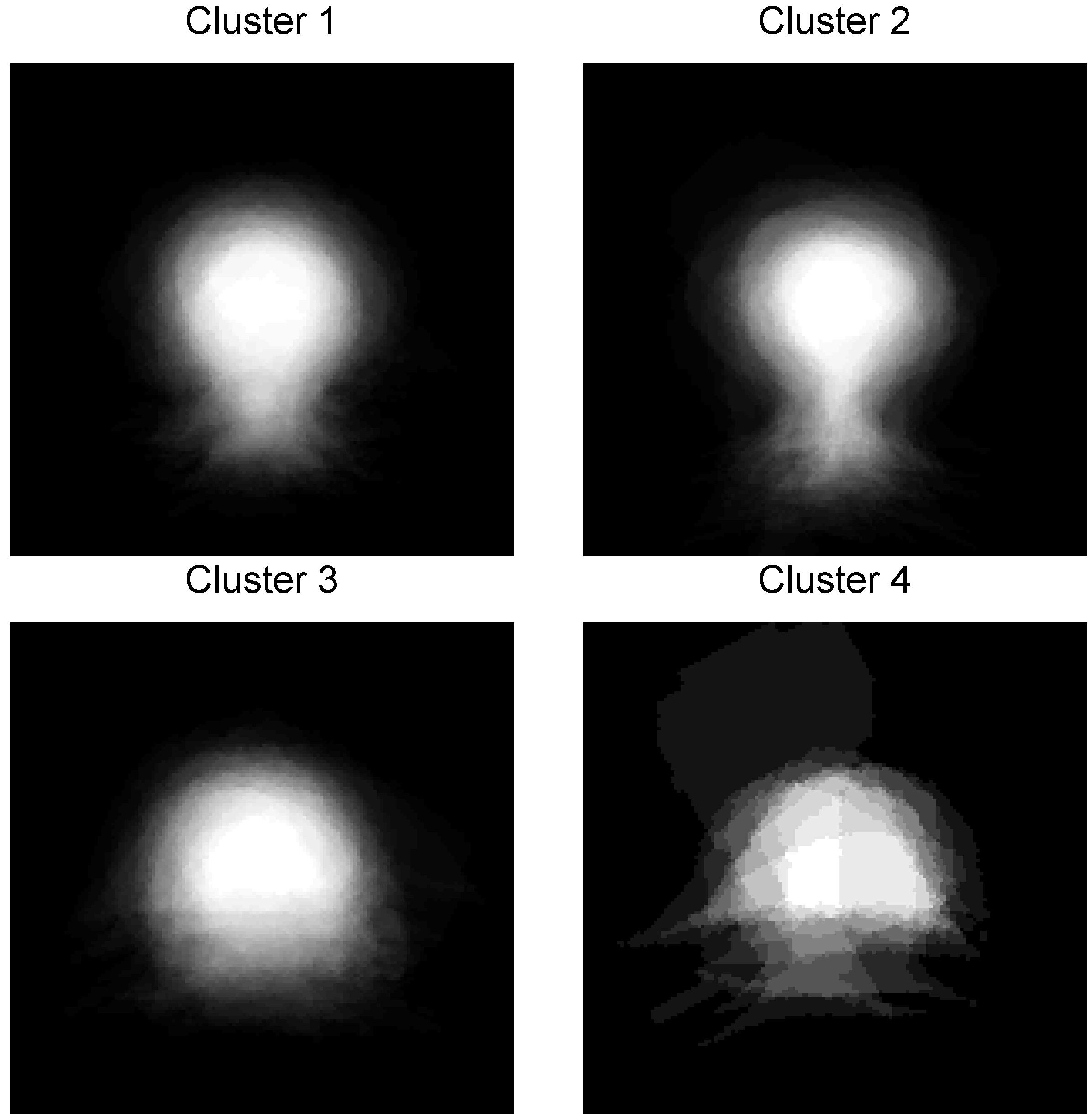}
	\caption{Average image for each cluster generated using morphological features.}
	\label{fig:cl_siu}
\end{figure}

\begin{figure}[tb]
	\centering
		\subfigure[Cluster 3]{
   \includegraphics[width=0.165\textwidth] {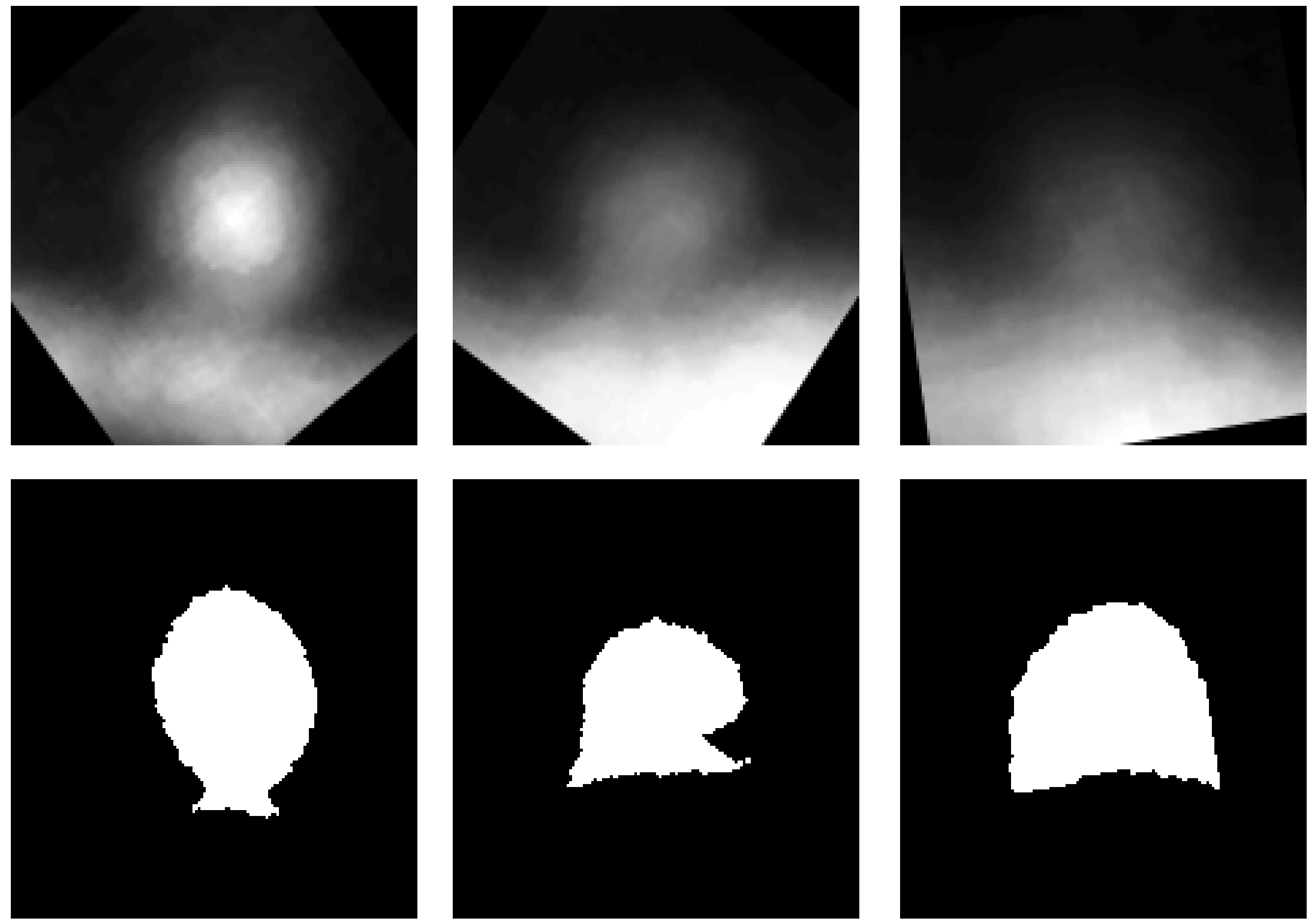}
   \label{fig:cl3_siu}
 }
 \hspace{30pt}
\subfigure[Cluster 4]{
   \includegraphics[width=0.18\textwidth] {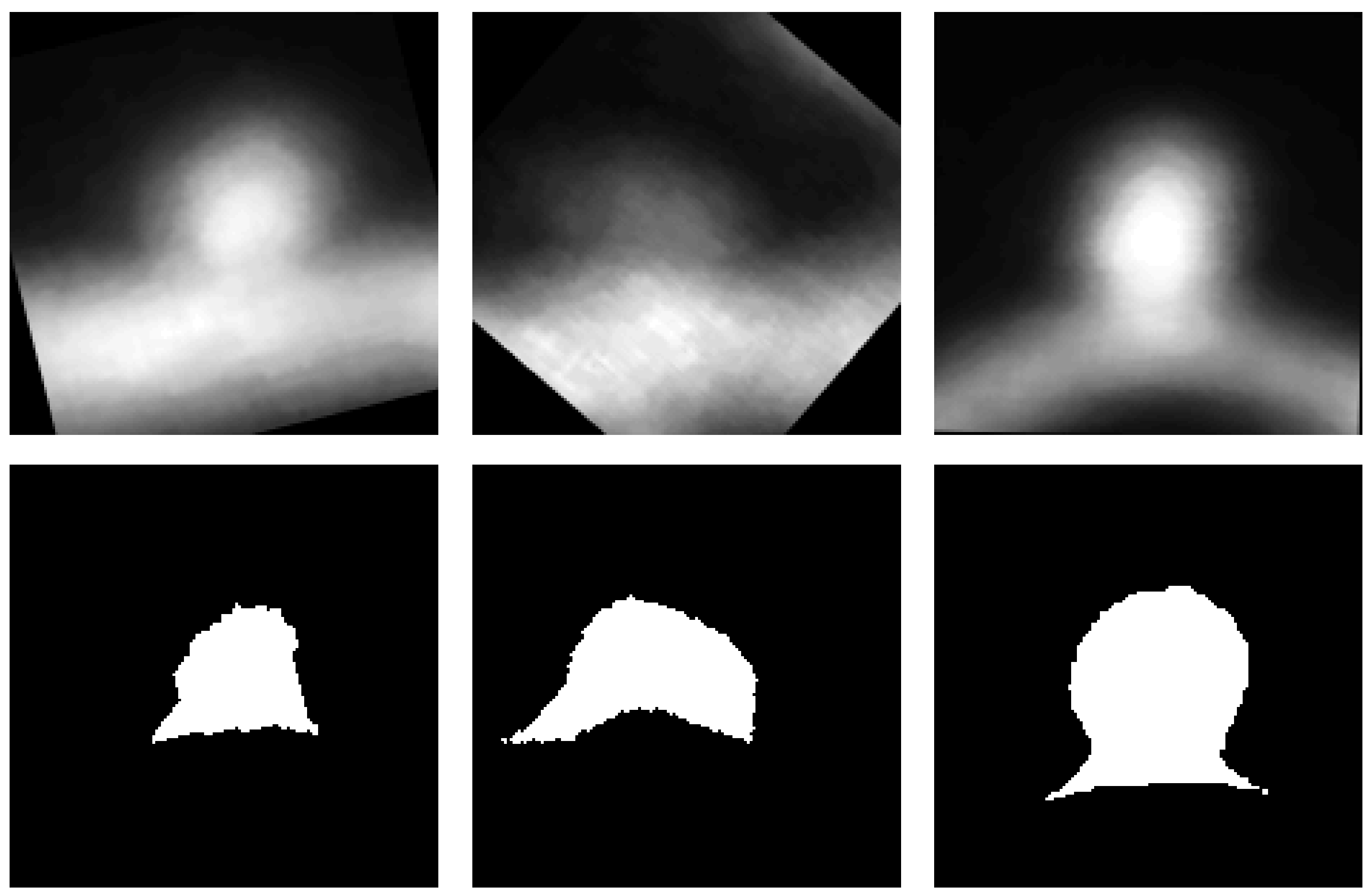}
   \label{fig:cl4_siu}
 }
	\caption{Intensity (top) and corresponding manually annotated images (bottom) for some of the spines from cluster $3$ and cluster $4$ using the morphology based features.}
	\label{fig:cl3_4_siu}
\end{figure}

Clustering analysis with morphological features resulted in $4$ clusters with sizes: $102$, $64$, $64$, and $12$ spines. Average image for each of the produced clusters is given in Fig. \ref{fig:cl_siu}. 
It is clear from Fig. \ref{fig:cl_siu} that cluster $1$, and $2$ are mushroom majority clusters. 
However, cluster $3$ and cluster $4$ show a mixed pattern, most of the spines have short thick neck, small head, and most importantly their neck diameters and head diameters are similar. A few spines from cluster $3$ and cluster $4$ along with their manually annotated images are presented in Fig. \ref{fig:cl3_4_siu}. These cluster appear to contain many stubby spines as well as spines with distinct heads and thick necks. 
It would be interesting to analyze which features are dominant in the clustering process, which might provide important information to neuroscientists. In this context, we perform an initial analysis using information gain \cite{infoGain91} and conclude that neck length is the most dominant feature for data used in this study, which confirms analysis performed in some of our previous studies \cite{Ghani_SIU15,ghani2016manifold}. \par

\subsection{Intensity profile features based analysis}
\begin{figure}[tb]
	\centering
		\includegraphics[width=0.35\textwidth]{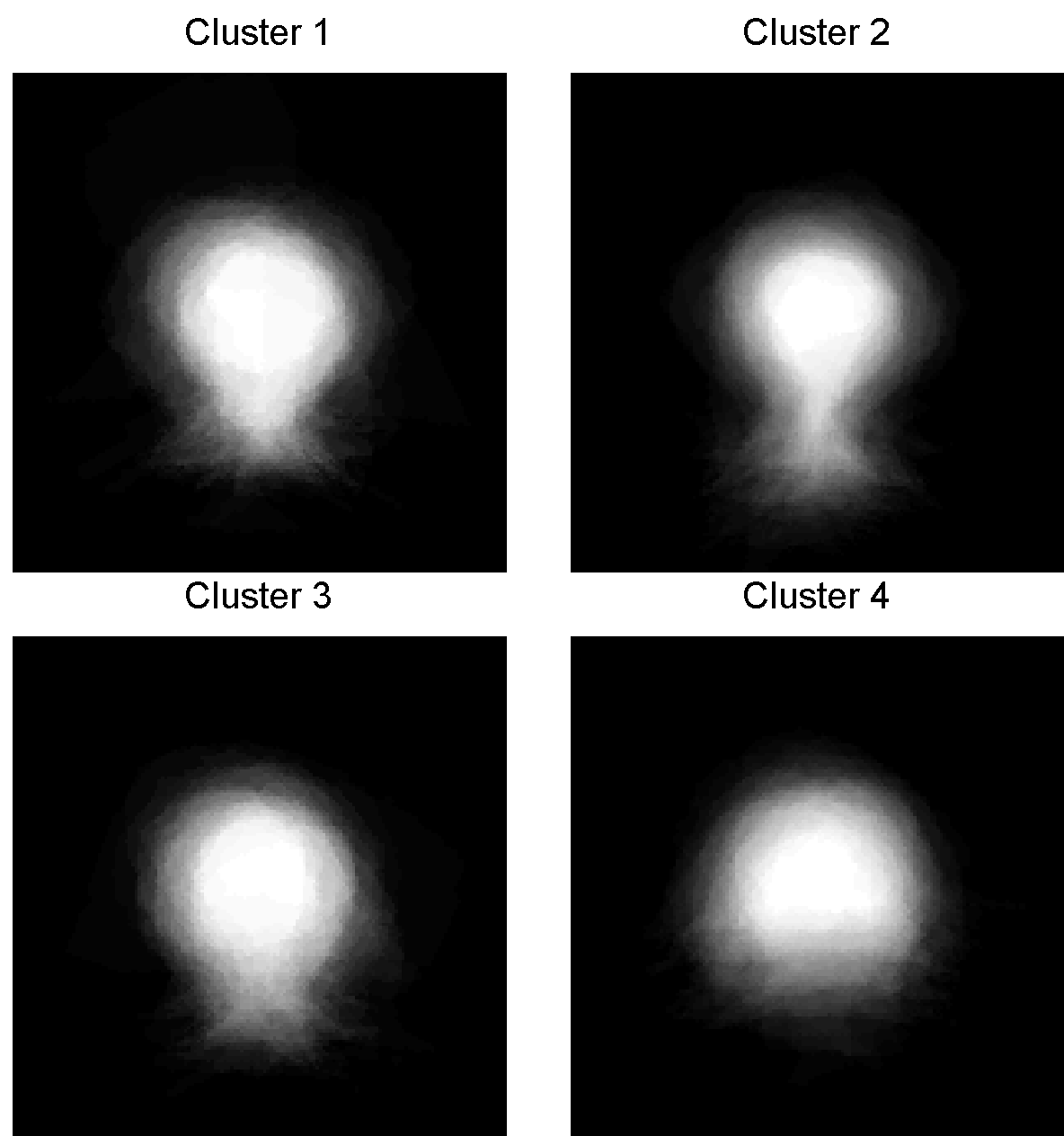}
	\caption{Average image for each cluster generated using the intensity profile based features.}
	\label{fig:cl_int}
\end{figure}

\begin{figure}[tb]
\centering
   \includegraphics[width=0.5\textwidth] {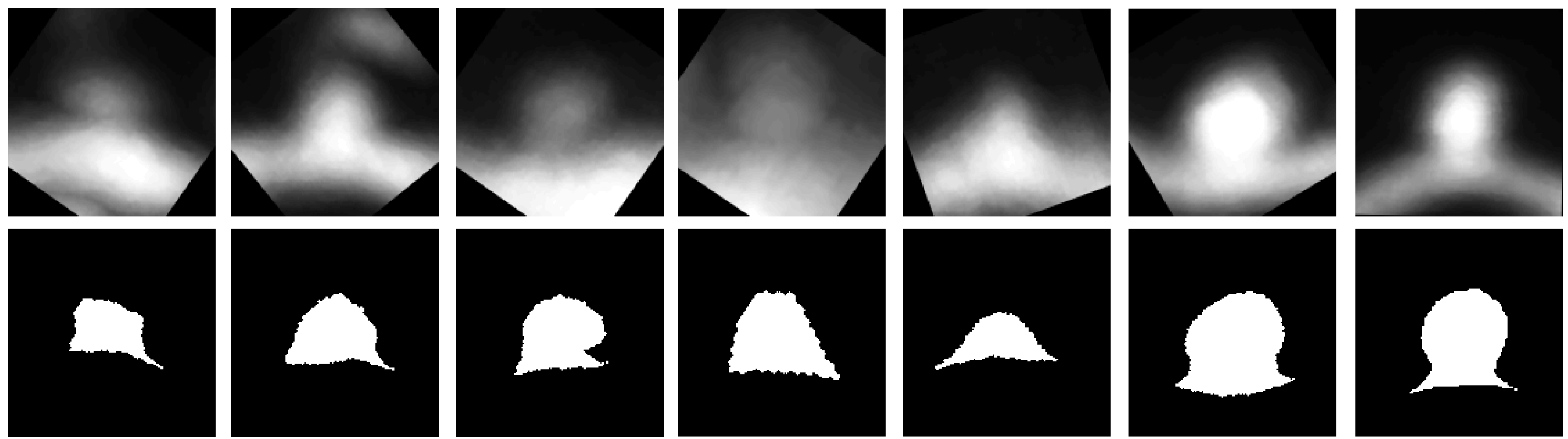}
\caption{Intensity (top) and corresponding manually annotated images (bottom) for some of the spines from cluster $4$ generated using the intensity profile based features.}
\label{fig:cl4_int}
\end{figure}

Using the intensity profile based features resulted in $4$ clusters consisting of $45$, $81$, $48$, and $68$ spines. The average image for each of these clusters is presented in Fig. \ref{fig:cl_int}. It is clear that cluster $1$, cluster $2$, and cluster $3$ are similar and appear to consist mostly of mushroom-like spines, i.e., they have big heads and long necks. Spines in cluster $3$ have relatively shorter necks as compared to cluster $2$, spines in cluster $4$ have big heads and very short or no necks. Some of the spines clustered in cluster $4$ are shown in Fig. \ref{fig:cl4_int}. 
This cluster appears to contain many stubby spines as well as spines with distinct heads and thick necks. \par

\subsection{Combined Features based Analysis}
\begin{figure}[tb]
	\centering
		\includegraphics[width=0.35\textwidth]{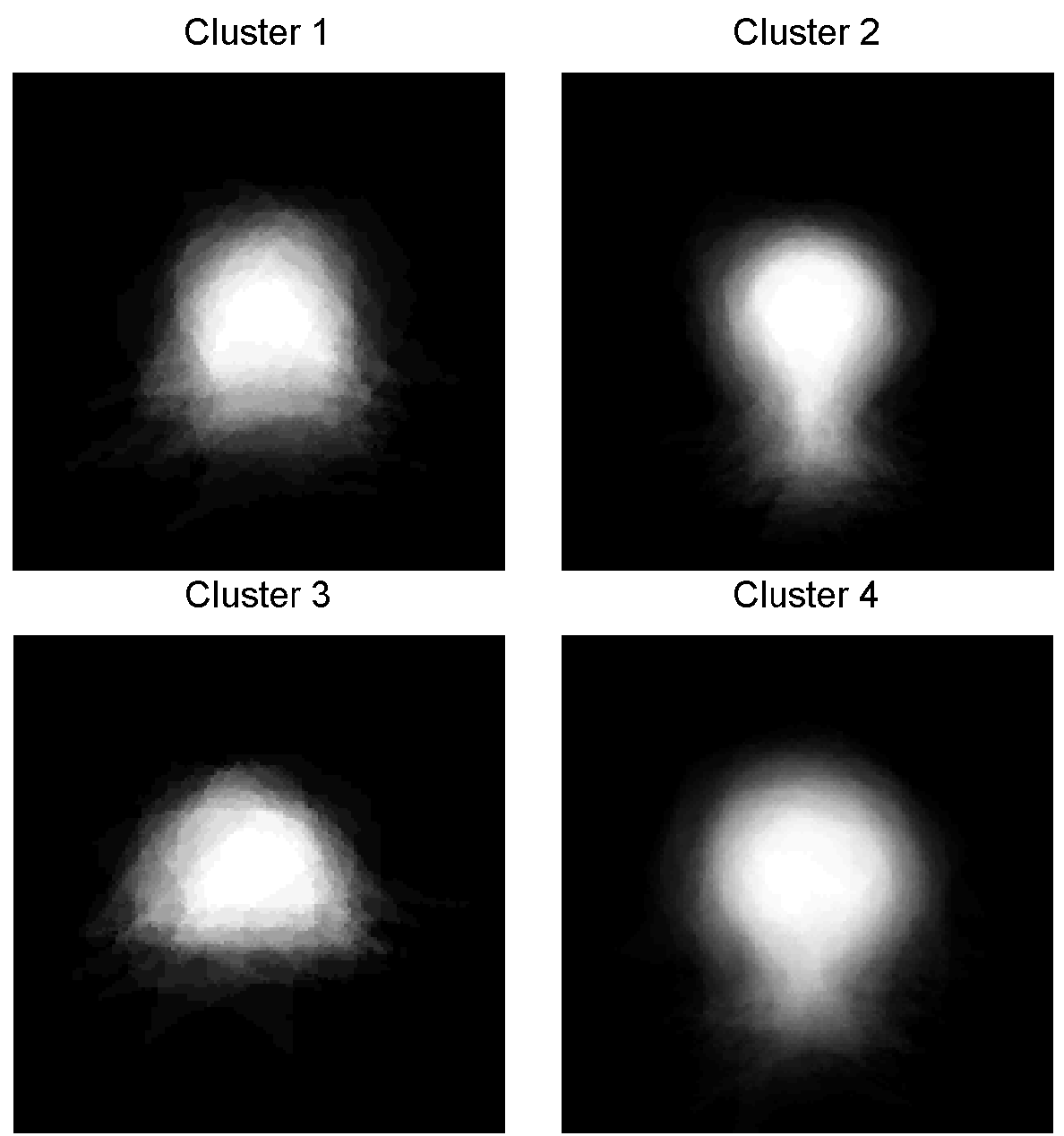}
	\caption{Average image for each cluster generated using HOG+DNSM features.}
	\label{fig:cl_hog_dnsm}
\end{figure}

\begin{figure}[tb]
	\centering
		\subfigure[Cluster 1]{
   \includegraphics[width=0.18\textwidth] {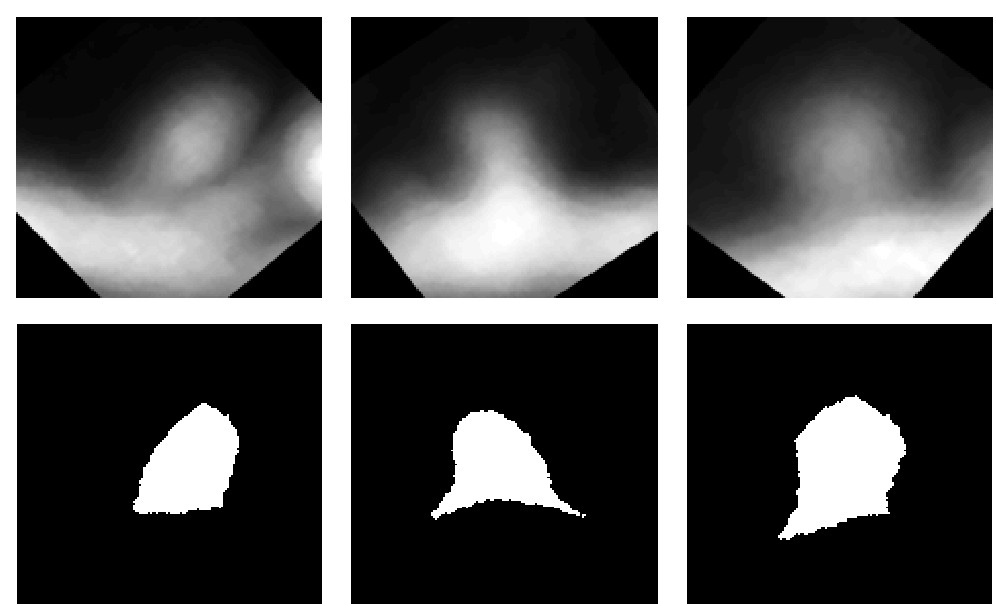}
   \label{fig:cl1_dnsm_hog}
 }
 \hspace{30pt}
\subfigure[Cluster 3]{
   \includegraphics[width=0.18\textwidth] {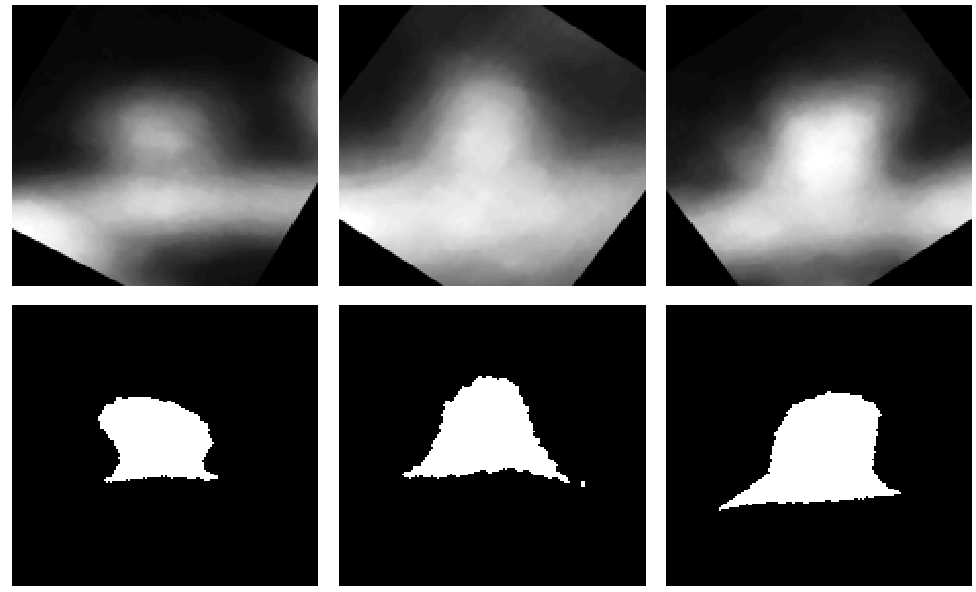}
   \label{fig:cl1_dnsm_hog}
 }
	\caption{Intensity (top) and corresponding manually annotated images (bottom) for some of the spines from cluster $1$ and cluster $3$ using HOG+DNSM based features.}
	\label{fig:cl1_3_hog_dnsm}
\end{figure}

Since, shape and appearance are complementary features, it is intuitive to combine both types of features and perform cluster analysis. 
We have already selected $100$ features from each group using a feature similarity based approach.
We combine these selected features to perform clustering in this section.
Using a combination of HOG and DNSM based features results in $4$ clusters consisting of $30$, $78$, $22$, and $112$ spines. The average image for each of these clusters is presented in Fig. \ref{fig:cl_hog_dnsm}. It is clear that cluster $2$ and cluster $4$ are similar and consist most of the mushroom-like spines, i.e., they have big heads and long necks. Spines in cluster $1$ and cluster $3$ are similar to one another in the sense that they have big heads and very short or no necks, as illustrated in Fig. \ref{fig:cl1_3_hog_dnsm}.

\begin{figure}[tb]
	\centering
		\includegraphics[width=0.35\textwidth]{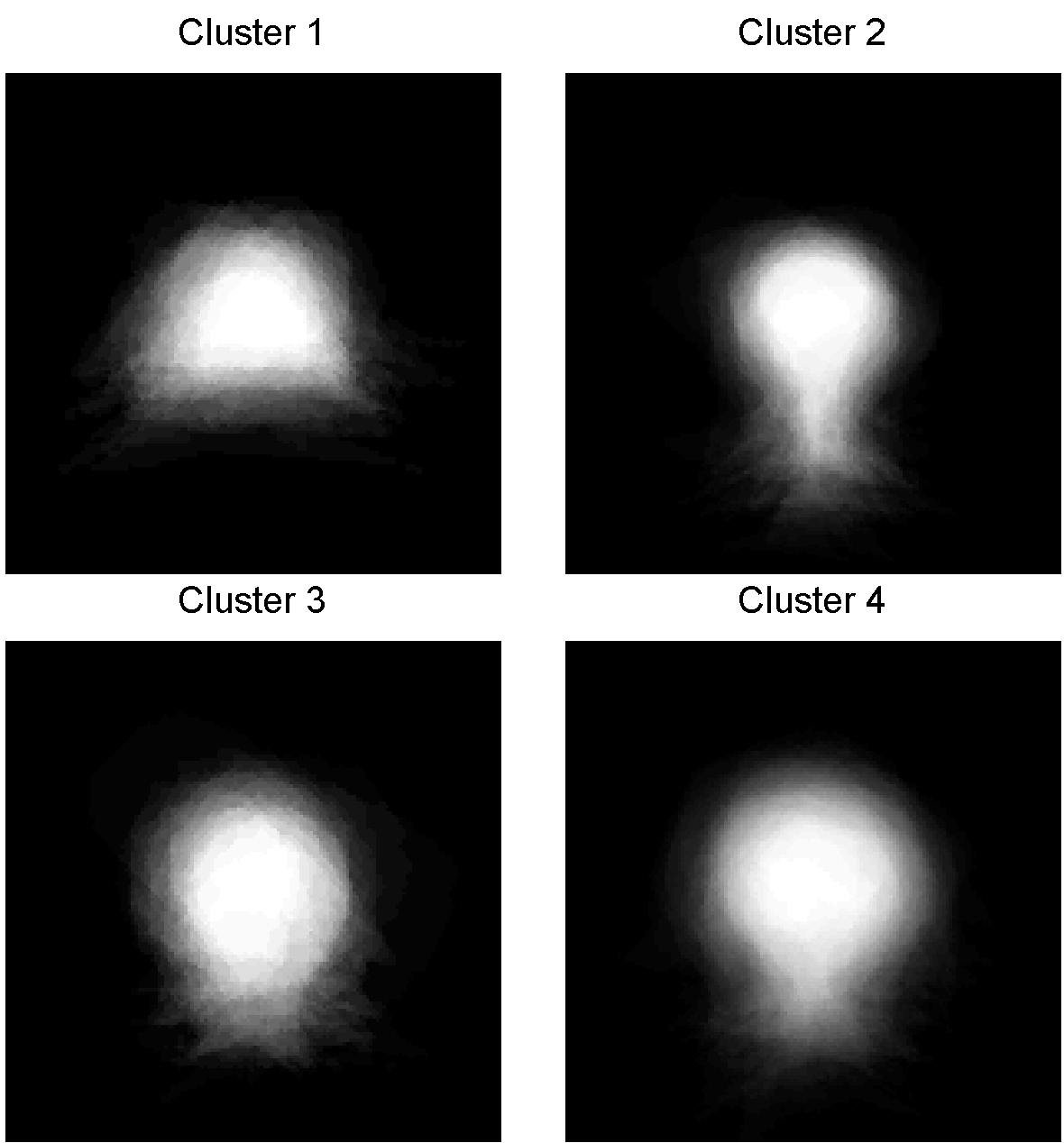}
	\caption{Average image for each cluster generated using DNSM+IntensityProfile features.}
	\label{fig:cl_dnsm_int}
\end{figure}

\begin{figure}[tb]
\centering
  \includegraphics[width=0.5\textwidth]{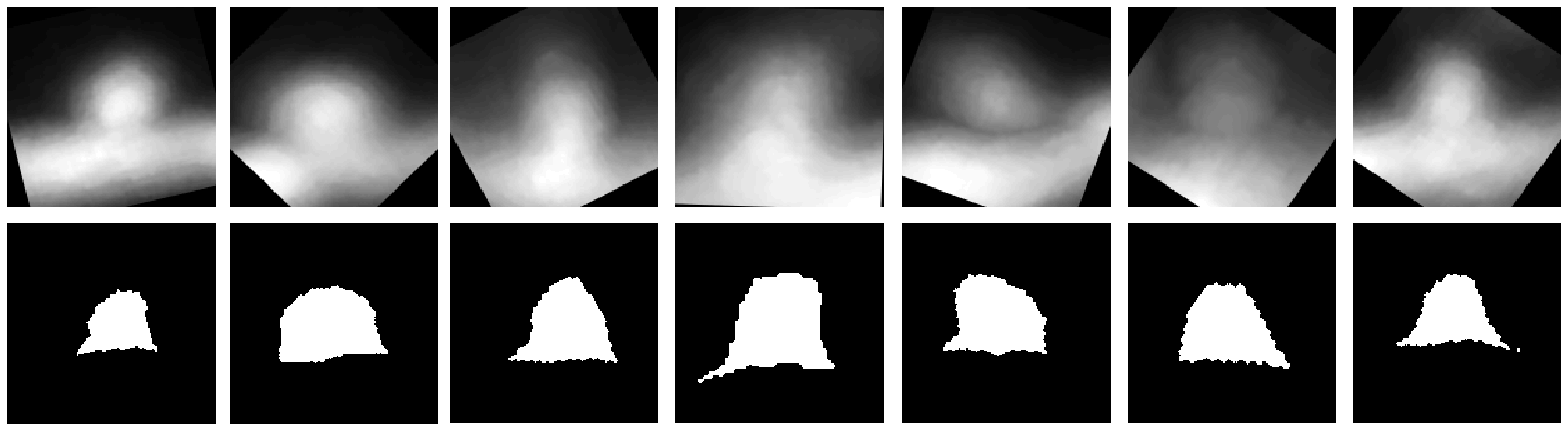}
\caption{Intensity (top) and corresponding manually annotated images (bottom) for some of the spines from cluster $1$ generated using DNSM+IntensityProfile features.}
\label{fig:cl1_dnsm_int}
\end{figure}

Using a combination of DNSM and intensity profile based features results in $4$ clusters consisting of $32$, $62$, $36$, and $112$ spines. Average image for each cluster is presented in Fig. \ref{fig:cl_dnsm_int}. Cluster $2$, cluster $3$, and cluster $4$ consist of mostly mushroom-like spines, having big heads and long necks. However, cluster $1$ consists of spines with intermediate properties: short, thick necks and big heads, as illustrated in Fig. \ref{fig:cl1_dnsm_int}. These spines have some morphological properties similar to mushroom spines and some similar to stubby spines, therefore, we may call cluster $1$ a mixed or intermediate cluster.  

\subsection{Clustering vs. Human Expert}
In this section, we compare the clustering results achieved using different representations to the labels assigned by a neuroscience expert. The idea is that similar data samples (belonging to same class) should be clustered in the same group. There are two challenges in spine shape analysis: (i) separating mushroom spines from stubby spines, and (ii) separating thin spines from filopodia type spines. Because of the developmental age of the animals we use, we see few filopodia in our data, this is why we focused on mushroom vs. stubby problem for this study. Stubby vs. mushroom analysis is a challenging task due to $2$PLSM resolution limits. In fact, in stimulated emission depletion (STED) microscopy images, many reported stubby spines look like mushroom spines \cite{tonnesen2014spine}. \par

\begin{table}[tb]
  \centering
  \caption{Comparison of clustering results and labels from human expert}
    \begin{tabular}{ccccccc}
    \hline
    \multirow{2}[2]{*}{Features} & \multirow{2}[2]{*}{Acc.} & \multirow{2}[2]{*}{Class} & \multicolumn{4}{c}{Clusters}  \\ 
          &           &  & 1     & 2     & 3     & 4 \\
\hline
    \multirow{2}[2]{*}{DNSM}  & \multirow{2}[2]{*}{79.34\%} & m & 11    & 48    & 38    & 85     \\
          &     & s       & 21    & 0     & 12     & 27     \\
          \hline
    \multirow{2}[2]{*}{Morphology}  & \multirow{2}[2]{*}{81.82\%} & m & 88   & 64    & 26    & 4     \\
          &      & s      & 14    & 0    & 38     & 8    \\
          \hline
    \multirow{2}[2]{*}{HOG}  & \multirow{2}[2]{*}{\textbf{88.02\%}} & m & 15    & 91    & 68    & 8      \\
          &   & s       & 34    & 2     & 4     & 20     \\
          \hline
          \multirow{2}[2]{*}{IntensityProfile}  & \multirow{2}[2]{*}{80.17\%} & m & 39    & 81    & 34    & 28      \\
          &   & s       & 6    & 0     & 14     & 40     \\
          \hline
                    \multirow{2}[2]{*}{HOG+DNSM}  & \multirow{2}[2]{*}{79.34\%} & m & 15    & 76    & 6    & 85      \\
          &   & s       & 15    & 2     & 16    & 27     \\
          \hline
                    \multirow{2}[2]{*}{DNSM+IntensityProfile}  & \multirow{2}[2]{*}{80.17\%} & m & 10    & 62    & 25    & 85      \\
          &   & s       & 22    & 0     & 11     & 27     \\
          \hline
    \end{tabular}%
  \label{tab:cl_exp}%
\end{table}%

A human expert manually labeled $242$ spine images, $182$ spines as mushroom and $60$ as stubby. Table \ref{tab:cl_exp} shows the class membership of the spines in each of the clusters formed using each feature type. We observe that some clusters are dominated by shapes from one class whereas other are mixed. We have already analyzed the similarity  within each of these clusters in the previous subsections, and observed the exploratory nature of our approach pointing to possibly intermediate shapes. Given the availability of manual labels, let us now carry out an analysis on the confirmatory aspects of our approach. In particular, to evaluate how strongly each clustering approach based on a different feature set confirms the manual shape labels, let us evaluate our clustering results using the manual labels as ground truth. To this end, let us pretend our clustering methods assign each cluster to the shape class with the majority of samples in that cluster. Then we can count the number of ``correct and incorrect classifications". Using this approach, we evaluate these feature representations and find out that HOG features perform best on the available data taking the human expert's labels as the ground truth, viewing this it as a classification problem we can achieve $88.02\%$ classification accuracy.\par

According to expert's labels, clusters $2$, $3$, and $4$ formed with the DNSM representation correspond to the mushroom class, whereas cluster $1$ is the stubby majority cluster. Sample images shown in Fig. \ref{fig:cl1_dnsm} suggests that spines in cluster $1$ have similar characteristics, however, the expert has labeled some of these spines as mushroom and others as stubby. This itself depicts the challenging nature of spine analysis and subjective nature of the manual labeling task. We have similar observations on clusters formed through the use of the other features. In particular, we observe both the confirmatory role of the clustering methods through the formation of clusters dominated by one of the classes as labeled by the human expert (e.g., HOG clusters $2$ and $3$), as well as the exploratory nature of clustering through the generation of clusters with mixed membership (e.g., HOG clusters $1$ and $4$). Our experimental analysis suggests that the possibility of intermediate shape types in addition to the conventional shape classes should be considered in spine shape analysis. One further step along this direction could involve efforts to characterize the distribution of spines in a continuous shape space.\par

\section{Conclusion} \label{conc}


In this paper, we have proposed a clustering approach to perform spine shape analysis. The advantages of adopting a clustering approach for spine shape analysis are: such an approach would not suffer from subjectivity, and analysis time would be reduced by avoiding manual labeling tasks. To the best of our knowledge an extensive clustering analysis of spine shapes has not been published. We use appearance, shape, and morphological feature based representations to perform clustering and shed some light on this problem. We perform clustering using x-means that uses BIC to select the number of clusters automatically; interestingly it produces $4$ clusters for all of the features considered here, this implies there are $4$ sub-groups in our data. Additionally, we have observed that, for the data used in our analysis, although there are many spines which easily fit into the definition of standard shape types (confirming the hypothesis), there are also a significant number of others which do not comply with standard shape types and demonstrate intermediate properties. Existence of intermediate shape types has been observed using all representations. It would be interesting to perform a neuroscientific analysis of produced clusters and understand biological meaning of each cluster produced. 
This is an initial analysis that provides clustering perspective on spine analysis and compare it with expert labels, it would also  be interesting to use proposed approach to perform an analysis tying clusters to different experimental conditions. \par

The emergence of this phenomenon can be explained in several ways. It is a known fact that dendritic spines exhibit shape type transitions over time, this phenomenon happens over the period of hours. If the spines are captured at these transition periods, for instance a mushroom spine changing to a stubby spine, it might happen to have a short and thick neck and a head diameter to neck diameter ratio close to $1$. As some spines in our data demonstrate such properties, it would be difficult to label them as mushroom or stubby. An alternative solution could be to define an intermediate class/group/cluster. A temporal analysis of several spine shapes would provide more insight into this phenomenon. It should also be noted that based on the expert labels, our data consists of two shape classes: mushroom and stubby. Including other shapes of spines such as thin and filopodia in the type of analysis we have proposed here might facilitate an even better understanding of the nature of shape classes and distribution. It might also be interesting to pose this as unsupervised regression problem, which would allow to study continuum of shape variations in a principled manner. It is important to mention that the distribution of spine shapes are dependent on various aspects of the data used, including which anatomical region of the brain the imaged neurons belong to as well as the age of the imaged neurons. This might also contribute towards different conclusions from different studies on spine shapes. Another potential issue might be performing the analysis on $2$D projections versus $3$D data. Therefore, it would be interesting to perform similar analysis with different $2$D projection methods as well as $3$D data. To conclude, the clustering perspective we propose in this paper can both be used to perform automated spine shape analysis to identify known shape classes as well as to help neuroscientists discover and explore unknown patterns in the shape space which have been previously ignored.\par

\section*{ACKNOWLEDGEMENT}
This work has been supported by the Scientific and Technological Research Council of Turkey (TUBITAK) under Grant $113E603$, and by TUBITAK-$2218$ Fellowship for Postdoctoral Researchers.

\clearpage
\bibliographystyle{splncs}
\bibliography{literature}

\end{document}